# DE-TGN: Uncertainty-Aware Human Motion Forecasting using Deep Ensembles

Kareem A. Eltouny, Wansong Liu, Sibo Tian, Minghui Zheng, *Member, IEEE*, and Xiao Liang, *Member, IEEE*

*Abstract*— Ensuring the safety of human workers in a collaborative environment with robots is of utmost importance. Although accurate pose prediction models can help prevent collisions between human workers and robots, they are still susceptible to critical errors. In this study, we propose a novel approach called deep ensembles of temporal graph neural networks (DE-TGN) that not only accurately forecast human motion but also provide a measure of prediction uncertainty. By leveraging deep ensembles and employing stochastic Monte-Carlo dropout sampling, we construct a volumetric field representing a range of potential future human poses based on covariance ellipsoids. To validate our framework, we conducted experiments using three motion capture datasets including Human3.6M, and two human-robot interaction scenarios, achieving state-of-the-art prediction error. Moreover, we discovered that deep ensembles not only enable us to quantify uncertainty but also improve the accuracy of our predictions.

## I. INTRODUCTION

The integration of automated robots into various industries has revolutionized repetitive task execution. As the demand for environmentally conscious manufacturing grows, there has been a surge in research on human-robot collaboration (HRC) to address electronic waste management tasks [1-3]. In an HRC environment, accurate human motion prediction plays a pivotal role in ensuring the safety of human workers. It empowers robots to anticipate human movement, enabling them to adjust their motion plans and avoid collisions [4]. Extensive studies have been conducted on 3D human motion forecasting, primarily leveraging motion capture technology. With the rapid advancements in artificial intelligence and its applications, machine learning methods have emerged for human motion prediction. These include recurrent neural networks (RNNs) [5-7], convolutional neural networks (CNNs) [8-11], graph convolutional networks (GCNs) [12-16], and transformers [17]. However, RNNs and transformers can be computationally demanding, while CNNs suffer from limited receptive fields, influenced by their kernel sizes. RNNs are also susceptible to pose discontinuities and error accumulation due to their step-by-step forecasting property.

Human motion is highly intricate, and accurately forecasting it entails dealing with a significant degree of uncertainty. In a collaborative robot setting, it is crucial for robots to recognize and account for such uncertain behaviors,

*Research supported by the National Science Foundation under Grant 2026533. (Corresponding authors: M. Zheng and X. Liang.)

K. A. Eltouny and X. Liang are with the Civil, Structural and Environmental Engineering Department, University at Buffalo, Buffalo, NY 14260 USA (e-mail: keltouny@buffalo.edu; liangx@buffalo.edu).

W. Liu, S. Tian, and M. Zheng are with the Mechanical and Aerospace Engineering Department, University at Buffalo, Buffalo, NY 14260 USA (e-mail: wansongl@buffalo.edu; sibotian@buffalo.edu; mhzheng@buffalo.edu).

allowing them to take appropriate actions when confidence level decreases. Several studies have been conducted to offer probabilistic outputs instead of deterministic ones, primarily through approximate variational inference and generative models [18-21]. However, traditional variational inference methods tend to generate samples from a local mode in the solution space, capturing only local uncertainty. Moreover, certain variational methods may impact prediction accuracy due to training constraints, such as imposing prior distributions on latent features in variational autoencoders. To address these limitations, deep ensembles have emerged as a potential solution. Deep ensembles encompass a collection of deep learning models that generate samples derived from distinct training trajectories [22]. By leveraging this ensemble approach, deep ensembles tackle the issue of local uncertainty by providing a broader range of potential predictions.

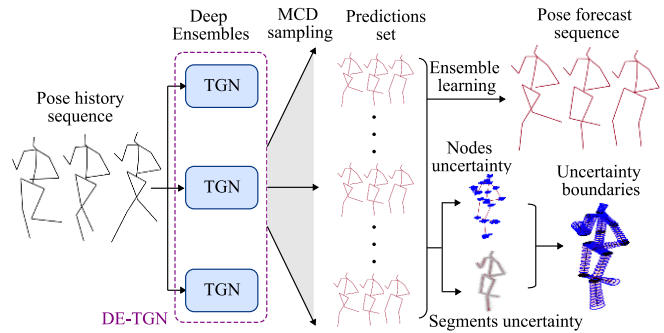

Figure 1. Overview of the proposed deep ensembles of temporal graph neural networks.

In this study, we introduce a novel approach called **D**eep **E**nsembles of **T**emporal-**G**raph Neural **N**etworks (DE-TGN) for accurate 3D human motion forecasting based on motion capture sequence data (Fig. 1). Our models employ a combination of temporal convolutional networks (TCN) and graph attention networks (GAT) to create a powerful hybrid architecture. We use deep ensembles in combination with Monte Carlo (MC) dropout sampling [23] to generate a diverse set of plausible motions. In addition, we propose a technique to construct 3D uncertainty boundaries using covariance ellipsoids derived from the probabilistic output. These boundaries provide valuable insights into the trustworthiness of the model's predictions in an HRC environment. Deep ensembles not only offer a diverse and representative set of solutions but also improve the quality and accuracy of the forecasts compared to using individual models. Additionally, TCNs rival the performance of other time-series modeling methods while benefiting from the efficiency of CNNs, making way for real-time applications. We also evaluated our method on Human3.6M [24], an established human motion prediction benchmark, and two human-robot interaction experiments.

## II. RELATED WORK

### A. Human motion prediction

In the past decade, the field of human motion forecasting has been dominated by RNNs, with several groundbreaking RNN-based methods proposed [5-7]. Fragkiadaki et al. [5] introduced one of the earliest RNN-based approaches, employing an encoder-decoder-RNN hybrid combined with curriculum learning for human motion forecasting. However, these RNN-based methods exhibited noticeable discontinuities at the beginning of the forecast. To address this issue, Martinez et al. [7] proposed a sequence-to-sequence model with residual connections which predicts velocities instead of poses. Despite these advancements, long-term predictions remain challenging for these methods due to their one-step-ahead prediction mode, leading to error accumulation and increased computational cost.

Feedforward networks, particularly CNNs, attempt to solve many of the inherited issues in RNN-based methods. Earlier methods, however, relied on the predefined human kinematic tree [8], overlooking the need for coordinated motion between the different body parts, even those that are distant [25]. In an effort to overcome these limitations, Li et al. [9] proposed a nested encoder-decoder CNN architecture for long-term motion forecasting. This approach involved convoluting over both the spatial and temporal axes, allowing for the capture of inter-joint spatial and temporal correlations. However, it is important to note that the temporal receptive field of CNNs is highly dependent on the kernel size. Additionally, treating the data as an image-like structure can pose challenges in effectively capturing the spatial correlations among joints.

### B. GCN

In recent years, there has been growing interest in using GCNs for human pose forecasting [12-16]. GCNs have shown promise in processing non-grid-like structures, such as the human pose, making them suitable for capturing inter-joint spatial correlations. Mao et al. [12] proposed a sequential, feed-forward network of GAT layers with fully connected graphs. This approach enables the learning of global spatial connectivity among joints through attention mechanisms in the trajectory space. In another study, Mao et al. [13] introduced motion attention layers to capture the similarity between the current motion and historical motion, resulting in more accurate predictions. To gain a deeper understanding of the spatiotemporal dynamics of joints, Sofianos et al. [15] proposed the use of depth-wise separable GCNs with trainable spatiotemporal adjacency matrices. Zhong et al. [16] took a mixture-of-experts approach in their GCN-based motion forecasting technique, where a gating network applies importance factors to a set of adjacency matrices.

### C. TCN

TCNs have gained attention as an efficient and effective alternative to RNN- and attention-based techniques for human motion forecasting, offering advantages such as reduced error accumulation and improved computational efficiency. However, the exploration of TCNs in this context has been limited compared to other time-series modeling methods. In a comparative study by Pavllo et al. [26], a GRU-based motion forecasting model was pitted against a WaveNet-based model, with the former demonstrating superior performance. Cui et al. [10] proposed a forecasting network consisting of GCN blocks that incorporated TCN layers to capture time dependencies. Li et al. [11] presented a similar approach but with the additional inclusion of a positional encoding module, allowing the network to predict action types alongside motion forecasting. Overall, despite the promising results and advantages offered by TCNs and dilated causal convolution in general, their applications in human motion forecasting remain relatively unexplored.

### D. Probabilistic learning

Several studies have put forth generative methods for human motion forecasting that aim to provide probabilistic output, allowing for diverse predictions without compromising accuracy. The rationale behind these approaches is the recognition that human motion forecasts should not be solely deterministic, particularly for long-term predictions [18]. Barsoum et al. [18] introduced HP-GAN, drawing inspiration from generative adversarial networks, which employs a sequence-to-sequence generator to predict a set of plausible human motion predictions. Aliakbarian et al. [19] noted the diversity of samples generated by HP-GAN decreases with training, as the network begins to disregard stochastic components. To address this issue, they proposed a recurrent-based conditional variational autoencoder (CVAE) with a mix-and-match strategy, randomly combining variation latent features with historic pose information. Another study by Yuan and Kitani [20] focused on diversifying generated samples and introduced the diversifying latent flows (DLow) sampling method, which utilizes a CVAE network. In a different approach, Salzmann et al. [21] proposed a typed graph-GRU hybrid to directly predict motion distributions, providing a probabilistic perspective. To the best of our knowledge, deep ensembles have not been investigated as a viable option for generating probabilistic output in human motion predictions.

## III. NETWORK ARCHITECTURE

In this section, we introduce the TGN architecture (Fig. 2) along with the Bayesian inference approximation using deep ensembles. Let us define $X_{1:N} = [x_1, x_2, ..., x_N]^T$ as the historical motion sequence consisting of $N$ 3D human poses. If the collaborating robot's motion is available, we denote its sequence of $N$ 3D poses as $Y_{1:N} = [y_1, y_2, ..., y_N]^T$. The vectors $x_i \in \mathbb{R}^{C_x}$ and $y_i \in \mathbb{R}^{C_y}$ contains $C_x$ and $C_y$ parameters, respectively, that describe the poses. The input to our network is the concatenation of these two sequences: $[X_{1:N}, Y_{1:N}]$. Our goal is to provide a forecast of the human motion poses for $T$ time steps, represented by the sequence $X_{N+1:N+T}$. We propose TGN, a TCN-GAT hybrid network, to predict the future sequence based on the provided input. To provide a measure of uncertainty, we rely on deep ensembles and MC dropout sampling to obtain a diverse set of predictions.

## A. GAT

GCNs are types of neural networks specifically designed to handle graph-structured data. Unlike CNNs, which operate on grid-like structured data with fixed local connectivity, GCNs consider varying connections for each node and its neighbors in the graph, as defined by an adjacency matrix. Among GCNs, GAT stands out as it utilizes self-attention mechanisms to assign varying importance to neighboring nodes, thereby adjusting the adjacency matrix [27]. In our model, we employ GAT to extract representative features that capture the spatial relations between pose nodes. Inspired by Mao et al. [12], we establish full connectivity among all nodes in the graph, allowing GAT to adapt the connections based on the available training data. In our case, we assume that both human and robot nodes form a fully-connected graph with a total of $C = C_x + C_y$ nodes. The edges of this graph can be represented by an adjacency matrix, denoted as $A \in \mathbb{R}^{C \times C}$. To transform the input into the trajectory space, we utilize the Discrete Cosine Transform (DCT). Consequently, each node is associated with a matrix $H \in \mathbb{R}^{C \times F}$, where $F$ represents the number of DCT coefficients. The graph convolutional layer estimates the output $H'$ using the following formula, acting as input for the subsequent layer:

$$H' = \sigma(A\,H\,W) \quad (1)$$

where $\sigma(\cdot)$ is an activation function and $W \in \mathbb{R}^{F \times F}$ is the trainable weight matrix. Using self-attention mechanisms, GAT applies attention weights to the entries in $A$ resulting in a learned adjacency matrix $A^*$ that replaces $A$ in Eq. (1):

$$A^* = \alpha \cdot A \quad (2)$$

where $\alpha \in \mathbb{R}^{C \times C}$ contains the edgewise attention weights obtained by averaging the output of multi-head attention.

To incorporate GAT layers into our model, we adopt a residual architecture, as depicted in Fig. 2b. Within a GAT block, two GAT layers are used, each followed by layer normalization [28], rectified linear unit (ReLU) activation [29], and dropout regularization [30]. A skip connection is then employed to merge the input with the output by means of element-wise addition. This residual architecture enables the blocks to focus on learning the relative changes in the feature maps, rather than the entire transformations, which can facilitate deep learning.

## B. TCN

TCNs are a specific type of CNNs designed to effectively handle sequential data [31, 32]. Unlike RNNs, TCNs do not rely on recurrent connections to capture the temporal dependencies. Instead, they employ dilated causal convolution on the input sequence. This allows for the easy attainment of large receptive fields, making TCNs capable of efficiently processing very long sequences while mitigating the risk of vanishing gradients. For a one-dimensional sequence input $x \in \mathbb{R}^N$ and a kernel $k: \{0, \dots, w-1\} \rightarrow \mathbb{R}$, the output $F(t)$ of a dilated convolution operation at step $t$ is defined by:

$$F(t) = \sum_{i=0}^{w-1} k(i) x(t - d \cdot i) + b \quad (3)$$

where $w$ represents the kernel size, $d$ is the dilation factor, and $b$ is the bias term. The dilation factor is often chosen with a base (e.g., 2) that doubles as the network gets deeper. Increasing the dilation factor $d$ and the kernel size $w$ in Equation (3) allows for the expansion of the receptive field.

Similar to the GAT residual blocks, the TCN modules in our model utilize a residual architecture, as illustrated in Fig. 2c. Each TCN block contains two sets of 1D dilated causal convolution layers, each followed by layer normalization, ReLU activation, and spatial dropout regularization [33]. Spatial dropout differs from traditional dropout by dropping entire channels, making it more suitable for nodes with high correlation to their neighborhood. Finally, the receptive field of $n$ successive TCN blocks can be defined as:

$$r = 1 + 2 \cdot (w - 1) \cdot \sum_{i=0}^{n-1} d^i \quad (4)$$

## C. Combined Architecture

The TGN architecture comprises three main modules: the GAT encoder, TCN encoder, and TCN decoder (Fig. 2a). The data undergoes DCT transformation and then passes through two GAT-Res blocks. Subsequently, the transformed data is converted back to the time domain using the inverse DCT operation. Both the TCN encoder and decoder consist of four TCN blocks, each corresponding to dilation values of 1, 2, 4, and 8. The TCN blocks utilize a kernel size of 3. The decoder concludes with a fully-connected layer that generates output corresponding to the number of desired forecasting steps. Additionally, a global residual connection is established between the last input step and the output, allowing the model's output to represent the relative position with respect to a query, which is typically the last known position [7].

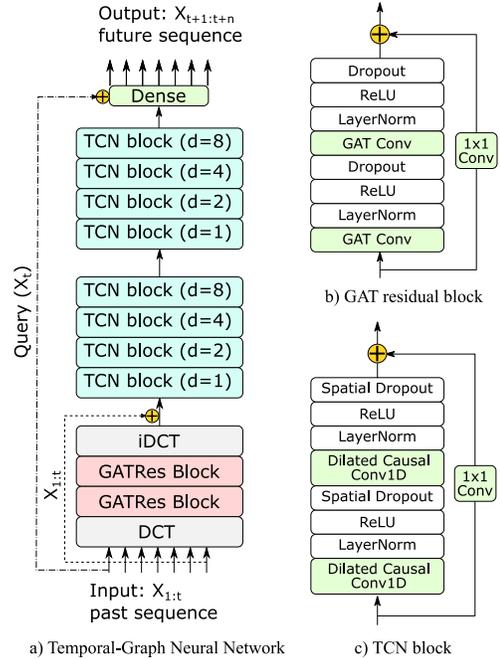

Figure 2. Temporal graph neural network architecture.

## D. Deep Ensembles

Deep ensemble, initially proposed by Lakshminarayanan et al. [22], is a machine learning approach that involves training multiple neural networks and combining their predictions to achieve improved accuracy. The concept of deep ensembles revolves around training several neural networks with different initializations, architectures, or training data. This ensemble approach helps mitigate the impact of random initialization and optimization on a single model's performance, resulting in enhanced predictions. It also aids in improving model robustness and uncertainty estimation. While deep ensembles were initially considered a "non-Bayesian" method for uncertainty quantification, there have been discussions about their approximation of the Bayesian posterior predictive distribution [34]. Nevertheless, deep ensembles offer advantages over standard Bayesian neural networks as they are easier to implement, require fewer computational resources, and involve minimal hyperparameter tuning. Deep ensembles have been shown to be effective across various applications, including image classification, natural language processing, and time-series forecasting [35].

A hypothesis suggests that deep ensembles perform well due to their ability to sample from unique functions or modes in the function (solution) space, as illustrated in Figure 3 [36]. In contrast, variational methods tend to focus on sampling from a single function, quantifying the uncertainty locally and potentially leading to a less diverse set of solutions. In addition, subsampling techniques may sample from a local optimum based on the training loss, but there is no guarantee that it corresponds to a local optimum of the validation loss.

In our model, we utilize deep ensembles to enhance prediction accuracy and quantify uncertainty. Specifically, we train three models with the same architecture but different parameter initializations. The predictions for future poses are obtained by averaging the node-wise outputs of these models. Although this may increase training time and memory requirements, there is minimal to no increase in inference time compared to variational inference methods. To quantify uncertainty, we combine deep ensembles with MC dropout sampling. This combination allows for an increased number of samples, enabling the construction of more robust distributions without significant computational overhead.

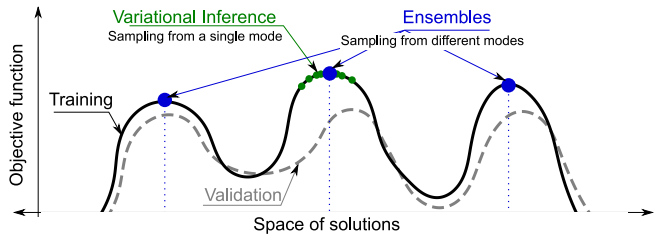

Figure 3. The unique functions sampling hypothesis.

## IV. UNCERTAINTY BOUNDARY

The stochastic output generated by the deep ensembles and the MC dropout sampling offers various possibilities for creating boundaries around pose estimates to indicate prediction uncertainty. In this section, we present an example of estimating uncertainty boundaries, which consists of two parts: 1) estimating uncertainty around the joints, and 2) estimating uncertainty along the segments.

### A. Joints Uncertainty

To establish an uncertainty boundary around the estimated joint position in an ensemble of predictions, we construct a covariance (error) ellipsoid. Each joint possesses three correlated dimensions. By performing an eigenvalue decomposition, we derive three principal axes that represent the joint's position, assuming they are uncorrelated. By treating the positional vector components along these axes as random variables, following a Gaussian distribution, we can construct a confidence boundary using a three-degree-of-freedom Chi-square distribution.

Given the global joint position vector $P = \{x, y, z\}^T$, the local position vector for the same joint is represented by $P' = \{x', y', z'\}^T$ where $x', y'$, and $z'$ are positions along the joint principal axes. The equation of the error ellipsoid in the local coordinates can be expressed as:

$$\left(\frac{x'}{\lambda_1}\right)^2 + \left(\frac{y'}{\lambda_2}\right)^2 + \left(\frac{z'}{\lambda_3}\right)^2 = \chi^2_{3,\alpha} \qquad (5)$$

where $\lambda_1, \lambda_2$, and $\lambda_3$ are the eigenvalues of the position vector ensemble, while $\chi^2_{3,\alpha}$ represents the third-degree Chi-square value at a significance level $\alpha$. To determine if a point falls outside the error ellipsoid, the left-hand side must be greater than the right-hand side (the critical Chi-square value). In global coordinates, the general formulas for the error ellipsoid are as follows:

$$\begin{bmatrix} x(\theta,\phi) \\ y(\theta,\phi) \\ z(\theta,\phi) \end{bmatrix} = \sqrt{\chi^2_{3,\alpha}} \cdot V\Lambda^{1/2} \begin{bmatrix} cos(\theta)\sin(\phi) \\ sin(\theta)\sin(\phi) \\ cos(\phi) \end{bmatrix} \qquad (6)$$

where $\theta$ and $\phi$ are the local azimuth and zenith, $V$ is the eigenvectors matrix, and $\Lambda^{1/2}$ is a diagonal matrix containing the square roots of the eigenvalues. The following general form can be used to determine if a point falls inside or outside the ellipsoid:

$$[x \quad y \quad z]\, V\Lambda^{-1/2}V^T \begin{bmatrix} x \\ y \\ Z \end{bmatrix} = \chi^2_{3,\alpha} \qquad (7)$$

where $\Lambda^{-1/2}$ represents a diagonal matrix containing the reciprocals of the square root of the eigenvalues.

### B. Segments Uncertainty

To represent body segments, we connect two joints according to the body kinematic tree. As our motion prediction model generates an ensemble of joint position predictions, connecting these joints results in a set of segment predictions. These segments can be described using line formulas of lines in 3D space. For a segment connecting two nodes, the model produces two groups of prediction points, one for each node. We construct the uncertainty boundary for the segment based on the mean segment connecting the mean points of the two groups, along with a dynamic 2D error ellipse that depends on the longitudinal position along the mean line.

There are various forms of the 3D line equation, including the symmetric form defined as:

$$\frac{x-x_0}{a} = \frac{y-y_0}{b} = \frac{z-z_0}{c} \qquad (8)$$

where $a$, $b$, and $c$ are the line parameters, and $p_0 = [x_0, y_0, z_0]$ is the line intersect. The parameters can be obtained given two points lying on the line: $p_1 = [x_1, y_1, z_1]$ and $p_2 = [x_2, y_2, z_2]$. The above formula can be rearranged in vector form, expressing $x$ and $y$ as functions of $z$:

$$\begin{Bmatrix} x(z) \\ y(z) \end{Bmatrix} = \begin{bmatrix} \beta_{11} & \beta_{12} \\ \beta_{21} & \beta_{22} \end{bmatrix} \begin{Bmatrix} z \\ 1 \end{Bmatrix} \quad (9)$$

where $\beta_{11} = \frac{x_2 - x_1}{z_2 - z_1}$, $\beta_{12} = x_1 - \left(\frac{x_2 - x_1}{z_2 - z_1}\right)z_1$, $\beta_{21} = \frac{y_2 - y_1}{z_2 - z_1}$, and $\beta_{22} = y_1 - \left(\frac{y_2 - y_1}{z_2 - z_1}\right)z_1$. We can define $P_h = [x(z), y(z)]^T$, $z_h = [z, 1]^T$, $\beta_1 = [\beta_{11}, \beta_{21}]^T$, $\beta_2 = [\beta_{12}, \beta_{22}]^T$, and then $\boldsymbol{B} = [\beta_1, \beta_2]$. Equation (9) can be rewritten as:

$$\boldsymbol{P}(z) = \boldsymbol{B} z_h \quad (10)$$

Since we have an ensemble of point pairs, $\boldsymbol{B}$ and consequently $\boldsymbol{P}(z)$, contain random variables and possess a variance that we exploit to construct the segment uncertainty boundary. Taking the variance of Eq. (10):

$$Var(\boldsymbol{P}) = z_h^T Var(\boldsymbol{B}) z_h \quad (11)$$

where

$$Var(\boldsymbol{B}) = \begin{bmatrix} Var(\beta_1) & Cov(\beta_1, \beta_2) \\ Cov(\beta_2, \beta_1) & Var(\beta_2) \end{bmatrix} \quad (12)$$

is a symmetric 4×4 matrix. $Var(\beta_1)$ and $Var(\beta_2)$ are regular covariance matrices while $Cov(\beta_1, \beta_2) = Cov(\beta_2, \beta_1)^T$ represents the cross-covariance matrices of the two random vectors $\beta_1$ and $\beta_2$. With some rearrangement, Eq. (11) becomes:

$$Var(\boldsymbol{P}) = z^2 Var(\beta_1) + z(Cov(\beta_1, \beta_2) + Cov(\beta_2, \beta_1)) + Var(\beta_2) \quad (13)$$

which is the 2×2 covariance matrix of the points on the lines intersecting with the $z$ plane. The covariance matrix can be used to construct a dynamic 2D error ellipse at any plane $z$ and is defined as follows in local coordinates:

$$\begin{bmatrix} x(t) \\ y(t) \end{bmatrix} = \sqrt{\chi_{\alpha,n}^2} \cdot \boldsymbol{V} \boldsymbol{\Lambda}^{1/2} \begin{bmatrix} \cos(t) \\ \sin(t) \end{bmatrix} \quad (14)$$

where $t \in [0, 2\pi]$. The general form for testing is:

$$[x \quad y] \cdot \boldsymbol{V} \boldsymbol{\Lambda}^{-1/2} \boldsymbol{V}^T \begin{bmatrix} x \\ y \end{bmatrix} = \chi_{\alpha,n}^2 \quad (15)$$

The following formulas can be used to transform the positional vectors of the points from global to local coordinates at the $z$ plane:

$$P_{local} = \boldsymbol{V}^{-1}(P_{global} - P_{origin}) \quad (16)$$

where $P_{global}$ is the point in the global coordinates, while $P_{origin}$ is the center of the error ellipse. The local axes are represented by three orthonormal dimensions, with the $z$-axis aligned with the mean segment's longitudinal dimension. Additionally, the local origin, $P_{origin}$, is chosen as the point on the mean segment at the plane of interest. The uncertainty boundary of the segments can be utilized to evaluate the proximity of robot segments by determining the points with the shortest distance between the robot and human segments.

## V. Experiments

To evaluate our model, we utilize three motion capture datasets: Human3.6M [24], the Arm Motion dataset [37], and the Reaching Motion dataset. We first provide an overview of these datasets, followed by details on the model implementation, evaluation metrics, and finally, the results.

### A. Datasets

*Human3.6M:* Human3.6M is a widely used publicly available dataset for motion capture data, particularly for human pose forecasting. It comprises motion capture recordings of seven actors performing 15 different actions, such as walking, eating, and engaging in discussions. Each pose includes the 3D Cartesian coordinates of 32 joints. We consider 17 joints after excluding joints with constant readings or close proximity to others. Following the approach in the literature [7], we use subject 5 for testing and subject 11 for validation. The remaining subjects (1, 6-9) are used for training. Additionally, we remove the global rotations and translation from each sequence and downsample all motions to 25 frames per second.

*Arm motion dataset:* This dataset focuses on the arm motion of human workers who grasp and relocate screwdrivers while being captured by the Vicon camera system. Only the trajectories of three nodes representing the arm motion are recorded. Three types of motions are performed, resulting in a total of 429 trajectories captured at a frequency of 25 Hz. The data is split into training, validation, and test sets using a ratio of 75/12.5/12.5, respectively, for each motion type.

*Reaching motion:* In this dataset, a human worker attempts to collect screws from different locations while a robot is moving in the shared space. This scenario represents an HRC environment and introduces complexities such as collision risks. The dataset includes 463 motion sequences recorded in 3D Cartesian coordinates, comprising six worker arm nodes and eight robotic arm nodes. The data is split into training, validation, and test sets using a ratio of 80/10/10, respectively.

### B. Evaluation Metrics

We employ the mean per joint position error (MPJPE) to assess prediction accuracy, measured in millimeters [24]. This metric is suitable for motion datasets represented in 3D Cartesian coordinates, unlike the more commonly used Euclidean distance for Euler angle representation [7, 12]. For a single future sequence $X_{N+1:N+T}$ and its corresponding prediction $\hat{X}$, the MPJPE value can be computed using the following formula:

$$E_{MPJPE}(X_{N+1:N+T}, \hat{X}_{N+1:N+T}) = \frac{1}{CT} \sum_{t=N+1}^{N+T} \sum_{c=1}^{C} \|\hat{x}_{c,t} - x_{c,t}\|^2 \quad (17)$$

where $C$, $N$, and $T$ represent the number of nodes, the number of historical frames, and the number of future frames, respectively. To evaluate the diversity of probabilistic predictions, we calculate the pairwise Euclidean distance between generated future poses based on the same historical motion[19]. Given a set of predictions $\hat{X}^i_{N+1:N+T}$, where $i = 1:S$ and $S$ represents the sample size, the diversity is estimated using the following formula:

$$Div = \frac{2}{S(S-1)} \sum_{i=1}^{S} \sum_{j=i+1}^{S} \frac{1}{T} \sum_{t=N+1}^{T} \|\hat{x}^i_t - \hat{x}^j_t\|^2 \quad (18)$$

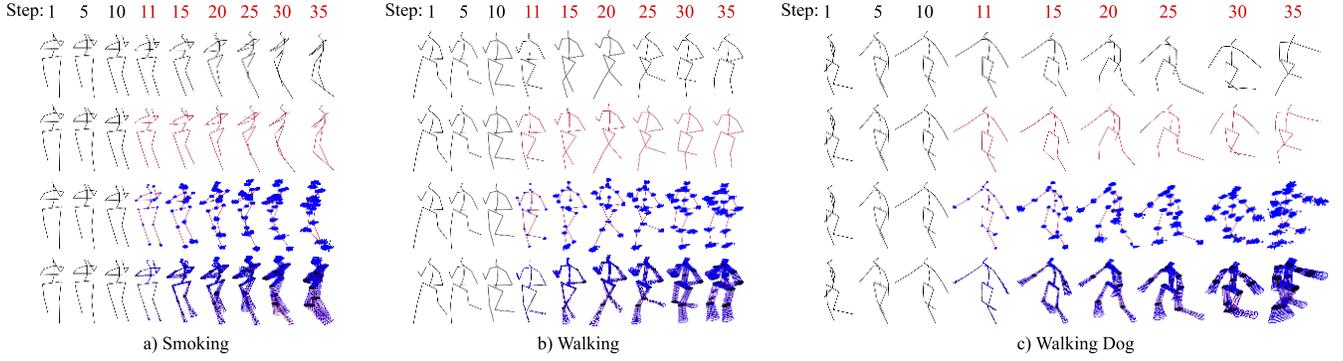

Figure 4. Examples of DE-TGN 10-25 (1000 milliseconds) predictions on Human3.6m including the uncertainty boundary. From top to bottom, we show the ground truth, DE-TGN mean predictions, DE-TGN + MC-dropout generated samples, constructed uncertainty boundary.

TABLE I. Human3.6M MPJPE values (mm) on the 15 action types at different forecasting steps for our proposed method (DE-TGN) trained on forecasting 10, 25, and 50 steps (400 ms, 1000 ms, and 2000 ms). Results of other methods in the literature are also provided (reported from [13] and [16]). *our models.

| | Directions | | | | | | Discussion | | | | | | Eating | | | | | | Greeting | | | | | |
|---|---|---|---|---|---|---|---|---|---|---|---|---|---|---|---|---|---|---|---|---|---|---|---|---|
| milliseconds | 80 | 160 | 320 | 400 | 1000 | 2000 | 80 | 160 | 320 | 400 | 1000 | 2000 | 80 | 160 | 320 | 400 | 1000 | 2000 | 80 | 160 | 320 | 400 | 1000 | 2000 |
| Res-GRU 25-25 [7] | 21.6 | 41.3 | 72.1 | 84.1 | 129 | | 25.7 | 47.8 | 80.0 | 91.3 | 132 | | 16.8 | 31.5 | 53.5 | 61.7 | 98.0 | | 31.2 | 58.4 | 96.3 | 109 | 154 | |
| ConvS2S 50-25 [9] | 13.5 | 29.0 | 57.6 | 69.7 | 116 | | 17.1 | 34.5 | 64.8 | 77.6 | 129 | | 11.0 | 22.4 | 40.7 | 48.4 | 87.1 | | 22.0 | 45.0 | 82.0 | 96.0 | 147 | |
| LTD 10-25 [12] | 9.2 | 20.6 | 46.9 | 58.8 | 109 | | 12.2 | 25.8 | 53.9 | 66.7 | 119 | | 7.7 | 15.8 | 30.5 | 37.6 | 74.1 | | 16.7 | 33.9 | 67.5 | 81.6 | 140 | |
| HRI 50-10 [13] | 7.4 | 18.4 | 44.5 | 56.5 | 107 | | 10.2 | 23.4 | 52.1 | 65.4 | 120 | | 7.0 | 14.9 | 29.9 | 36.4 | 75.7 | | 13.7 | 30.1 | 67.8 | 78.1 | 139 | |
| GAGCN 10-25[16] | 7.3 | 12.8 | 30.3 | 34.5 | 69.9 | | 9.7 | 17.1 | 31.4 | 38.9 | 76.9 | | 6.4 | 11.5 | 21.7 | 25.2 | 51.4 | | 11.8 | 20.1 | 40.5 | 48.4 | 87.7 | |
| DE-TGN 10-10* | 3.1 | 3.8 | 4.0 | 5.9 | | | 3.6 | 4.9 | 5.4 | 7.1 | | | 3.8 | 4.9 | 5.2 | 7.4 | | | 3.8 | 4.2 | 4.3 | 5.8 | | |
| DE-TGN 10-25* | 5.9 | 8.6 | 8.3 | 8.2 | 12.9 | | 5.5 | 8.6 | 10.3 | 10.2 | 14.9 | | 5.9 | 8.9 | 10.3 | 10.5 | 15.4 | | 6.0 | 8.7 | 9.0 | 9.0 | 13.5 | |
| DE-TGN 10-50* | 8.0 | 12.9 | 15.4 | 15.4 | 13.8 | 19.4 | 7.6 | 12.7 | 15.7 | 16.6 | 16.0 | 23.1 | 7.6 | 12.7 | 15.7 | 16.0 | 18.4 | 25.6 | 8.8 | 14.3 | 15.5 | 14.9 | 16.4 | 20.8 |
| | Phoning | | | | | | Posing | | | | | | Purchases | | | | | | Sitting | | | | | |
| milliseconds | 80 | 160 | 320 | 400 | 1000 | 2000 | 80 | 160 | 320 | 400 | 1000 | 2000 | 80 | 160 | 320 | 400 | 1000 | 2000 | 80 | 160 | 320 | 400 | 1000 | 2000 |
| Res-GRU 25-25 [7] | 21.1 | 38.9 | 66.0 | 76.4 | 126 | | 29.3 | 56.1 | 98.3 | 114 | 183 | | 28.7 | 52.4 | 86.9 | 1001 | 154 | | 23.8 | 44.7 | 78.0 | 91.2 | 153 | |
| ConvS2S 50-25 [9] | 13.5 | 26.6 | 49.9 | 59.9 | 114 | | 16.9 | 36.7 | 75.7 | 92.9 | 187 | | 20.3 | 41.8 | 76.5 | 89.9 | 152 | | 13.5 | 27.0 | 52.0 | 63.1 | 121 | |
| LTD 10-25 [12] | 10.2 | 20.2 | 40.9 | 50.9 | 105 | | 12.5 | 27.5 | 62.5 | 79.6 | 172 | | 15.5 | 32.3 | 63.6 | 77.3 | 136 | | 10.4 | 21.4 | 45.4 | 57.3 | 119 | |
| HRI 50-10 [13] | 8.6 | 18.3 | 39.0 | 49.2 | 105 | | 10.2 | 24.2 | 58.5 | 75.8 | 178 | | 13.0 | 29.2 | 60.4 | 73.9 | 134 | | 9.3 | 20.1 | 44.3 | 56.0 | 116 | |
| GAGCN 10-25[16] | 8.8 | 13.5 | 25.5 | 28.7 | 66.0 | | 10.1 | 17.0 | 35.5 | 45.1 | 99.1 | | 11.9 | 20.7 | 41.8 | 47.6 | 85.1 | | 9.3 | 14.4 | 29.6 | 38.5 | 71.1 | |
| DE-TGN 10-10* | 4.0 | 5.2 | 5.5 | 7.3 | | | 3.4 | 3.9 | 4.0 | 5.4 | | | 3.7 | 4.3 | 4.4 | 6.0 | | | 3.4 | 4.6 | 5.0 | 6.9 | | |
| DE-TGN 10-25* | 6.4 | 9.5 | 10.7 | 10.9 | 16.4 | | 6.1 | 8.1 | 8.5 | 8.5 | 12.9 | | 6.0 | 8.4 | 8.3 | 8.3 | 13.2 | | 5.3 | 8.1 | 9.7 | 9.8 | 15.3 | |
| DE-TGN 10-50* | 8.6 | 14.2 | 17.1 | 17.3 | 19.0 | 29.4 | 8.9 | 14.1 | 15.0 | 14.7 | 14.4 | 17.3 | 10.8 | 15.9 | 16.6 | 17.3 | 16.4 | 19.3 | 7.6 | 12.4 | 16.6 | 17.0 | 17.0 | 24.4 |
| | Sitting Down | | | | | | Smoking | | | | | | Taking Photo | | | | | | Waiting | | | | | |
| milliseconds | 80 | 160 | 320 | 400 | 1000 | 2000 | 80 | 160 | 320 | 400 | 1000 | 2000 | 80 | 160 | 320 | 400 | 1000 | 2000 | 80 | 160 | 320 | 400 | 1000 | 2000 |
| Res-GRU 25-25 [7] | 31.7 | 58.3 | 96.7 | 112 | 187 | | 18.9 | 34.7 | 57.5 | 65.4 | 102 | | 21.9 | 41.4 | 74.0 | 87.6 | 154 | | 23.8 | 44.2 | 75.8 | 87.7 | 135 | |
| ConvS2S 50-25 [9] | 20.7 | 40.6 | 70.4 | 82.7 | 150 | | 11.6 | 22.8 | 41.3 | 48.9 | 81.7 | | 12.7 | 26.0 | 52.1 | 63.6 | 128 | | 14.6 | 29.7 | 58.1 | 69.7 | 118 | |
| LTD 10-25 [12] | 17.0 | 33.4 | 61.6 | 74.4 | 144 | | 8.4 | 16.8 | 32.5 | 39.5 | 75.6 | | 9.9 | 20.5 | 43.8 | 55.2 | 120 | | 10.5 | 21.6 | 45.9 | 57.1 | 107 | |
| HRI 50-10 [13] | 14.9 | 30.7 | 59.1 | 72.0 | 144 | | 7.0 | 14.9 | 29.9 | 36.4 | 69.5 | | 8.3 | 18.4 | 40.7 | 51.5 | 116 | | 8.7 | 19.2 | 43.4 | 54.9 | 108 | |
| GAGCN 10-25[16] | 14.1 | 24.8 | 40.0 | 47.4 | 84.1 | | 7.1 | 11.8 | 21.7 | 24.3 | 48.7 | | 8.5 | 13.9 | 28.8 | 35.1 | 70.0 | | 8.5 | 14.1 | 29.8 | 33.8 | 69.3 | |
| DE-TGN 10-10* | 3.8 | 5.1 | 5.5 | 7.5 | | | 3.3 | 4.5 | 4.9 | 6.7 | | | 3.0 | 3.7 | 4.0 | 5.4 | | | 3.3 | 4.2 | 4.4 | 6.1 | | |
| DE-TGN 10-25* | 5.7 | 9.0 | 10.2 | 10.5 | 17.3 | | 5.4 | 8.2 | 9.6 | 9.7 | 14.6 | | 5.3 | 7.5 | 7.9 | 7.8 | 11.5 | | 5.5 | 8.1 | 8.8 | 8.7 | 13.7 | |
| DE-TGN 10-50* | 8.4 | 14.0 | 17.7 | 18.4 | 18.1 | 26.8 | 7.2 | 11.8 | 14.6 | 14.8 | 17.1 | 23.8 | 7.4 | 12.0 | 13.5 | 13.0 | 13.0 | 21.1 | 8.2 | 13.7 | 15.9 | 15.4 | 16.3 | 23.6 |
| | Walking | | | | | | Walking Dog | | | | | | Walk Together | | | | | | Average | | | | | |
| milliseconds | 80 | 160 | 320 | 400 | 1000 | 2000 | 80 | 160 | 320 | 400 | 1000 | 2000 | 80 | 160 | 320 | 400 | 1000 | 2000 | 80 | 160 | 320 | 400 | 1000 | 2000 |
| Res-GRU 25-25 [7] | 23.2 | 40.9 | 61.0 | 66.1 | 79.1 | | 36.4 | 64.8 | 99.1 | 111 | 166 | | 20.4 | 37.1 | 59.4 | 67.3 | 98.2 | | 25.0 | 46.2 | 77.0 | 88.3 | 137 | |
| ConvS2S 50-25 [9] | 17.7 | 33.5 | 56.3 | 63.6 | 82.3 | | 27.7 | 53.6 | 90.7 | 103 | 162 | | 15.3 | 30.4 | 53.1 | 61.2 | 87.4 | | 16.6 | 33.3 | 61.4 | 72.7 | 124 | |
| LTD 10-25 [12] | 12.6 | 23.6 | 39.4 | 44.5 | 60.9 | | 22.9 | 43.5 | 74.5 | 86.4 | 142 | | 10.8 | 21.7 | 39.6 | 47.0 | 65.7 | | 12.4 | 25.2 | 49.9 | 60.9 | 113 | |
| HRI 50-10 [13] | 10.0 | 19.5 | 34.2 | 39.8 | 58.1 | | 20.1 | 40.3 | 73.3 | 86.3 | 147 | | 8.9 | 18.4 | 35.1 | 41.9 | 69.6 | | 10.4 | 22.6 | 47.1 | 58.3 | 112 | |
| GAGCN 10-25[16] | 10.3 | 16.1 | 28.8 | 23.4 | 51.1 | | 17.0 | 28.8 | 50.1 | 59.4 | 91.3 | | 8.8 | 13.8 | 26.2 | 29.9 | 50.4 | | 10.1 | 16.9 | 32.5 | 38.5 | 77.3 | |
| DE-TGN 10-10* | 5.4 | 6.9 | 7.9 | 10.6 | | | 5.7 | 6.5 | 7.1 | 9.3 | | | 4.2 | 5.0 | 5.8 | 7.7 | | | 3.9 | 5.0 | 5.4 | 7.3 | | |
| DE-TGN 10-25* | 8.7 | 12.6 | 13.8 | 14.0 | 21.6 | | 9.2 | 12.8 | 13.3 | 13.9 | 21.8 | | 7.3 | 10.3 | 11.0 | 11.2 | 17.1 | | 6.3 | 9.3 | 10.3 | 10.4 | 16.0 | |
| DE-TGN 10-50* | 12.2 | 19.9 | 23.1 | 22.7 | 24.2 | 37.0 | 14.3 | 23.0 | 21.9 | 21.6 | 23.7 | 33.8 | 9.9 | 15.6 | 16.5 | 15.9 | 18.2 | 27.4 | 8.9 | 14.5 | 17.1 | 17.2 | 18.0 | 26.0 |

## C. Implementation

To perform training and testing, all motions are divided into fixed-length windows, which serve as observations. For Human3.6M and Arm Motion datasets, three separate experiments are conducted, each with a different output length. In all experiments, the input sequences (history) consist of 10 steps (equivalent to 400 milliseconds), while the output sequences (forecasts) are 10, 25, and 50 steps long, respectively. Additionally, two experiments are carried out for the reaching motion dataset, using 10-step input sequences and output sequences of 25 and 50 steps. Two more experiments are conducted for the Reaching Motion dataset, incorporating robot motion in the input sequences. All models are built and trained using TensorFlow [38] and the Adam optimizer [39]. The total number of trainable parameters varies between 2.99E6 and 3.68E6, depending on the sequence length and the number of nodes. The models are trained in batches of 32 for 400 epochs or until convergence is achieved.

## D. Results

*Human3.6M:* Table I presents the MPJPE values for the three DE-TGN models. trained on the Human3.6M dataset. The values are provided for each of the 15 actions and different forecasting steps. Additionally, results from other models in the literature are included for comparison. Our proposed DE-TGN models outperform all other models across all actions. However, it is worth noting that models trained to produce long-term forecasts perform worse in shorter-term predictions compared to models focused on short-term forecasts (e.g., DE-TGN 10-50 vs. DE-TGN 10-10). As a result, there are a few instances where other models outperform our long-term model (DE-TGN 10-50) in the 80 milliseconds forecast range (e.g., Walk Together). This trade-off indicates that long-term forecast models sacrifice short-term forecast accuracy to achieve exceptional accuracy in long-term predictions. This observation is supported by significant improvements in long-term predictions when compared to state-of-the-art models. Fig. 4 showcases examples of DE-TGN predictions and the estimation of uncertainty boundary at multiple time steps.

Table II presents the average MPJPE values for each individually trained model in the deep ensembles, covering the 10, 25, and 50 steps variants. Notably, the deep ensembles technique achieves lower MPJPE values compared to all individual TGN models used to construct DE-TGN. Similarly, Table III demonstrates that deep ensembles provide increased diversity, as evidenced by higher APD values compared to all individual models. It should be noted that 32 samples are used for each individual model, while 33 samples are utilized for deep ensembles to ensure a fair comparison.

TABLE II. Human3.6M average MPJPE values (mm) over all actions for individual TGN models and their deep ensembles.

| Forecast length | 400 ms | 1000 ms | 2000 ms |
|---|---|---|---|
| TGN #1 | 6.24 | 12.7 | 20.8 |
| TGN #2 | 6.65 | 12.7 | 23.2 |
| TGN #3 | 6.18 | 13.9 | 20.7 |
| DE-TGN | **5.02** | **10.6** | **17.7** |

TABLE III. Human3.6M average APD values (mm) over all actions for individual TGNs with MC dropout sampling and their deep ensembles.

| Forecast length | 400 ms | 1000 ms | 2000 ms |
|---|---|---|---|
| TGN #1 MC-dropout | 38.51 | 78.33 | 126.48 |
| TGN #2 MC-dropout | 40.48 | 79.97 | 131.19 |
| TGN #3 MC-dropout | 38.06 | 80.88 | 125.16 |
| DE-TGN + MC-dropout | **45.66** | **92.14** | **147.11** |

*Arm Motion Dataset:* The MPJPE results for the Arm Motion dataset are displayed in Table IV. For comparison, we include the results of a residual sequence-to-sequence (Seq2Seq) GRU-based model with input and output lengths of 25 steps each. The DE-TGN 10-25 variant not only outperforms the Seq2Seq model but also utilizes a shorter input length and requires fewer computational resources due to the efficiency of the convolutional layers. Moreover, the DE-TGN 10-50 variant achieves slightly improved predictions compared to the Seq2Seq model while offering double the forecast length. The results also demonstrate that deep ensembles reduce prediction errors in all models.

TABLE IV. Arm Motion Dataset average MPJPE values (mm) over all actions a for individual models and their deep ensembles.

| Model type | TGN 10-10 | TGN 10-25 | TGN 10-50 | Seq2Seq 25-25 |
|---|---|---|---|---|
| Model #1 | 1.94 | 4.21 | 7.83 | 8.27 |
| Model #2 | 1.89 | 4.23 | 7.99 | 8.17 |
| Model #3 | 1.91 | 4.18 | 7.99 | 8.24 |
| Deep Ensembles | 1.83 | 4.06 | 7.61 | 7.89 |

*Reaching Motion Dataset:* Table V presents the MPJPE values for all individual TGN models and their ensembles in the Reaching Motion dataset. Similar observations to the previous experiments are noted, such as improvements in prediction accuracy due to the utilization of the deep ensembles technique and an increase in errors as the forecast length grows. Additionally, Table V examines the effects of including robot motion in the input sequence. While there is no significant impact on error values when including robot motion in the 25-step forecast models, we observe a decrease in error for the 50-step forecast models, suggesting that the forecasting model may benefit from incorporating robot motion in long-term predictions. Fig. 5 showcases example predictions on the reaching motion dataset.

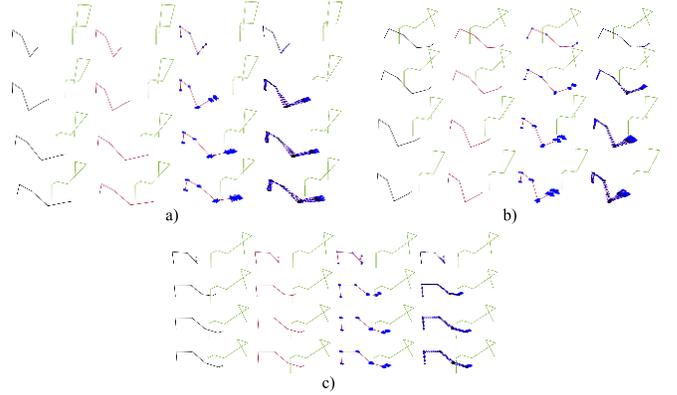

Figure 5. Three Examples (a-c) of DE-TGN predictions for the Reaching Motion Dataset at time steps 11, 30, 50, and 60. From left to right, we show the ground truth, mean predictions, generated samples, and constructed uncertainty boundary. The robot arm is shown in green.

TABLE V. Reaching Motion Dataset average MPJPE values (mm) for TGNs and their deep ensembles, with and without robot motion input.

| Model type | 10-25 | 10-25 w/o robot input | 10-50 | 10-50 w/o robot input |
|---|---|---|---|---|
| TGN #1 | 7.70 | 7.51 | 17.67 | 17.84 |
| TGN #2 | 7.30 | 7.50 | 18.23 | 19.36 |
| TGN #3 | 7.67 | 7.59 | 17.44 | 19.05 |
| DE-TGN | **7.18** | **7.22** | **16.62** | **17.54** |

## VI. CONCLUSION

This study highlights the advantages of employing deep ensembles for human motion forecasting. We have introduced the DE-TGN architecture, which surpasses the current state-of-the-art methods in human motion prediction for the Human3.6M benchmark, while also offering longer-term forecasts. Furthermore, our models have demonstrated low prediction errors in two HRC datasets that capture the motions of human workers engaged in collaborative tasks with a robotic arm. We have also proposed a statistical method for estimating uncertainty boundaries of human body nodes and segments utilizing deep ensembles and MC dropout sampling.

Leveraging convolutional layers, our approach proves to be highly efficient compared to traditional sequence-to-sequence models. By providing accurate predictions and assessing the reliability of the models through uncertainty estimation, our framework lays a solid foundation for safer HRC. In future studies, we will further investigate the practical effectiveness of the estimated forecasting uncertainty boundaries in the context of HRC.

REFERENCES

[1] M.-L. Lee, W. Liu, S. Behdad, X. Liang, and M. Zheng, "Robot-assisted disassembly sequence planning with real-time human motion prediction," *IEEE Transactions on Systems, Man, and Cybernetics: Systems,* vol. 53, no. 1, pp. 438-450, 2022.
[2] S. Sajedi, W. Liu, K. Eltouny, S. Behdad, M. Zheng, and X. Liang, "Uncertainty-assisted image-processing for human-robot close collaboration," *IEEE Robotics and Automation Letters,* vol. 7, no. 2, pp. 4236-4243, 2022.
[3] X. Zhang, K. Eltouny, X. Liang, and S. Behdad, "Automatic Screw Detection and Tool Recommendation System for Robotic Disassembly," *Journal of Manufacturing Science and Engineering,* vol. 145, no. 3, p. 031008, 2023.
[4] W. Liu, X. Liang, and M. Zheng, "Task-Constrained Motion Planning Considering Uncertainty-Informed Human Motion Prediction for Human–Robot Collaborative Disassembly," *IEEE/ASME Transactions on Mechatronics,* 2023.
[5] K. Fragkiadaki, S. Levine, P. Felsen, and J. Malik, "Recurrent network models for human dynamics," in *Proceedings of the IEEE international conference on computer vision*, 2015, pp. 4346-4354.
[6] A. Jain, A. R. Zamir, S. Savarese, and A. Saxena, "Structural-rnn: Deep learning on spatio-temporal graphs," in *Proceedings of the ieee conference on computer vision and pattern recognition*, 2016, pp. 5308-5317.
[7] J. Martinez, M. J. Black, and J. Romero, "On human motion prediction using recurrent neural networks," in *Proceedings of the IEEE conference on computer vision and pattern recognition*, 2017, pp. 2891-2900.
[8] J. Butepage, M. J. Black, D. Kragic, and H. Kjellstrom, "Deep representation learning for human motion prediction and classification," in *Proceedings of the IEEE conference on computer vision and pattern recognition*, 2017, pp. 6158-6166.
[9] C. Li, Z. Zhang, W. S. Lee, and G. H. Lee, "Convolutional sequence to sequence model for human dynamics," in *Proceedings of the IEEE conference on computer vision and pattern recognition*, 2018, pp. 5226-5234.
[10] Q. Cui, H. Sun, and F. Yang, "Learning dynamic relationships for 3d human motion prediction," in *Proceedings of the IEEE/CVF conference on computer vision and pattern recognition*, 2020, pp. 6519-6527.
[11] B. Li, J. Tian, Z. Zhang, H. Feng, and X. Li, "Multitask non-autoregressive model for human motion prediction," *IEEE Transactions on Image Processing,* vol. 30, pp. 2562-2574, 2020.
[12] W. Mao, M. Liu, M. Salzmann, and H. Li, "Learning trajectory dependencies for human motion prediction," in *Proceedings of the IEEE/CVF International Conference on Computer Vision*, 2019, pp. 9489-9497.
[13] W. Mao, M. Liu, and M. Salzmann, "History repeats itself: Human motion prediction via motion attention," in *Computer Vision–ECCV 2020: 16th European Conference, Glasgow, UK, August 23–28, 2020, Proceedings, Part XIV 16*, 2020: Springer, pp. 474-489.
[14] M. Li, S. Chen, Y. Zhao, Y. Zhang, Y. Wang, and Q. Tian, "Dynamic multiscale graph neural networks for 3d skeleton based human motion prediction," in *Proceedings of the IEEE/CVF conference on computer vision and pattern recognition*, 2020, pp. 214-223.
[15] T. Sofianos, A. Sampieri, L. Franco, and F. Galasso, "Space-time-separable graph convolutional network for pose forecasting," in *Proceedings of the IEEE/CVF International Conference on Computer Vision*, 2021, pp. 11209-11218.
[16] C. Zhong, L. Hu, Z. Zhang, Y. Ye, and S. Xia, "Spatio-temporal gating-adjacency GCN for human motion prediction," in *Proceedings of the IEEE/CVF Conference on Computer Vision and Pattern Recognition*, 2022, pp. 6447-6456.
[17] E. Aksan, M. Kaufmann, P. Cao, and O. Hilliges, "A spatio-temporal transformer for 3d human motion prediction," in *2021 International Conference on 3D Vision (3DV)*, 2021: IEEE, pp. 565-574.
[18] E. Barsoum, J. Kender, and Z. Liu, "Hp-gan: Probabilistic 3d human motion prediction via gan," in *Proceedings of the IEEE conference on computer vision and pattern recognition workshops*, 2018, pp. 1418-1427.
[19] S. Aliakbarian, F. S. Saleh, M. Salzmann, L. Petersson, and S. Gould, "A stochastic conditioning scheme for diverse human motion prediction," in *Proceedings of the IEEE/CVF Conference on Computer Vision and Pattern Recognition*, 2020, pp. 5223-5232.
[20] Y. Yuan and K. Kitani, "Dlow: Diversifying latent flows for diverse human motion prediction," in *Computer Vision–ECCV 2020: 16th European Conference, Glasgow, UK, August 23–28, 2020, Proceedings, Part IX 16*, 2020: Springer, pp. 346-364.
[21] T. Salzmann, M. Pavone, and M. Ryll, "Motron: Multimodal probabilistic human motion forecasting," in *Proceedings of the IEEE/CVF Conference on Computer Vision and Pattern Recognition*, 2022, pp. 6457-6466.
[22] B. Lakshminarayanan, A. Pritzel, and C. Blundell, "Simple and scalable predictive uncertainty estimation using deep ensembles," *Advances in neural information processing systems,* vol. 30, 2017.
[23] Y. Gal and Z. Ghahramani, "Dropout as a bayesian approximation: Representing model uncertainty in deep learning," in *international conference on machine learning*, 2016: PMLR, pp. 1050-1059.
[24] C. Ionescu, D. Papava, V. Olaru, and C. Sminchisescu, "Human3. 6m: Large scale datasets and predictive methods for 3d human sensing in natural environments," *IEEE transactions on pattern analysis and machine intelligence,* vol. 36, no. 7, pp. 1325-1339, 2013.
[25] D. A. Winter, "Human balance and posture control during standing and walking," *Gait & posture,* vol. 3, no. 4, pp. 193-214, 1995.
[26] D. Pavllo, C. Feichtenhofer, M. Auli, and D. Grangier, "Modeling human motion with quaternion-based neural networks," *International Journal of Computer Vision,* vol. 128, pp. 855-872, 2020.
[27] P. Veličković, G. Cucurull, A. Casanova, A. Romero, P. Liò, and Y. Bengio, "Graph Attention Networks," in *International Conference on Learning Representation*, 2018.
[28] J. L. Ba, J. R. Kiros, and G. E. Hinton, "Layer Normalization," in *Neural Information Processing Systems (NIPS)*, 2016.
[29] V. Nair and G. E. Hinton, "Rectified linear units improve restricted boltzmann machines," in *The 27th international conference on machine learning (ICML-10)*, 2010, pp. 807-814.
[30] N. Srivastava, G. Hinton, A. Krizhevsky, I. Sutskever, and R. Salakhutdinov, "Dropout: a simple way to prevent neural networks from overfitting," *The journal of machine learning research,* vol. 15, no. 1, pp. 1929-1958, 2014.
[31] A. v. d. Oord *et al.*, "Wavenet: A generative model for raw audio," *arXiv preprint arXiv:1609.03499,* 2016.
[32] S. Bai, J. Z. Kolter, and V. Koltun, "An empirical evaluation of generic convolutional and recurrent networks for sequence modeling," *arXiv preprint arXiv:1803.01271,* 2018.
[33] J. Tompson, R. Goroshin, A. Jain, Y. LeCun, and C. Bregler, "Efficient object localization using convolutional networks," in *Proceedings of the IEEE conference on computer vision and pattern recognition*, 2015, pp. 648-656.
[34] A. G. Wilson and P. Izmailov, "Bayesian deep learning and a probabilistic perspective of generalization," *Advances in neural information processing systems,* vol. 33, pp. 4697-4708, 2020.
[35] Y. Ovadia *et al.*, "Can you trust your model's uncertainty? evaluating predictive uncertainty under dataset shift," *Advances in neural information processing systems,* vol. 32, 2019.
[36] S. Fort, H. Hu, and B. Lakshminarayanan, "Deep ensembles: A loss landscape perspective," *arXiv preprint arXiv:1912.02757,* 2019.
[37] W. Liu, X. Liang, and M. Zheng, "Dynamic model informed human motion prediction based on unscented kalman filter," *IEEE/ASME Transactions on Mechatronics,* vol. 27, no. 6, pp. 5287-5295, 2022.
[38] M. Abadi *et al.*, "Tensorflow: Large-scale machine learning on heterogeneous distributed systems," *arXiv preprint arXiv:1603.04467,* 2016.
[39] D. P. Kingma and J. Ba, "Adam: A method for stochastic optimization," in *The International Conference on Learning Representations (ICLR)*, 2015.